\documentclass[10pt,twocolumn,letterpaper]{article}

\usepackage{cvpr}
\usepackage{algorithm}
\usepackage{algpseudocode}
\usepackage{times}
\usepackage{epsfig}
\usepackage{graphics}
\usepackage{amsmath}
\usepackage{amssymb}
\usepackage{ulem}
\usepackage{subcaption,graphicx}
\usepackage{color}
\usepackage{mathrsfs}
\usepackage[table]{xcolor}
\usepackage{booktabs}
\usepackage{makecell}
\usepackage{colortbl}
\usepackage{multirow}
\usepackage{array}
\usepackage{colortbl}
\definecolor{mygray1}{gray}{.9}
\definecolor{mygray2}{gray}{.5}

\usepackage[pagebackref=true,breaklinks=true,letterpaper=true,colorlinks,bookmarks=false]{hyperref}

\definecolor{mygray1}{gray}{.9}
\definecolor{mygray2}{gray}{.5}

{}
{}
{}
\cvprfinalcopy 

\def\cvprPaperID{216} 

\ifcvprfinal\fi
\begin{document}

\title{Deep Transfer Across Domains for Face Anti-spoofing}

\author{\normalsize{Xiaoguang~Tu$^{1}$\footnote{Contact Author}, Hengsheng~Zhang$^{1}$, Mei~Xie$^{1}$, Yao~Luo$^{1}$, Yuefei~Zhang$^{2}$, Zheng~Ma$^{1}$} \\
	\small{$^{1}$University of Electronic Science and Technology of China, $^{2}$Chongqing Institute of Public Security Science and Technology} \\
	{\small  xguangtu@outlook.com}}
\maketitle

\begin{abstract}
A practical face recognition system demands not only high recognition performance, but also the capability of detecting spoofing attacks. While emerging approaches of face anti-spoofing have been proposed in recent years, most of them do not generalize well to new database. The generalization ability of face anti-spoofing needs to be significantly improved before they can be adopted by practical application systems. The main reason for the poor generalization of current approaches is the variety of materials among the spoofing devices. As the attacks are produced by putting a spoofing display (e.t., paper, electronic screen, forged mask) in front of a camera, the variety of spoofing materials can make the spoofing attacks quite different. Furthermore, the background/lighting condition of a new environment can make both the real accesses and spoofing attacks different. Another reason for the poor generalization is that limited labeled data is available for training in face anti-spoofing. In this paper, we focus on improving the generalization ability across different kinds of datasets. We propose a CNN framework using sparsely labeled data from the target domain to learn features that are invariant across domains for face anti-spoofing. Experiments on public-domain face spoofing databases show that the proposed method significantly improve the cross-dataset testing performance only with a small number of
labeled samples from the target domain.
\end{abstract}


\section{Introduction}
In biometric based face recognition systems, spoofing attacks are usually perpetrated using photographs, replayed videos or forged masks. Despite continuous works on face anti-spoofing over the
years \cite{Pereira2012LBP,Yang2014Learn,Pinto2015Using,li2016original,tu2019enhance,tu2019learning}, most of them limit to the non-realistic intra-database testing scenarios instead of the cross database testing scenarios. In practical scenario, the environments are not fixed, differences in light conditions, backgrounds and camera resolutions of a new environment may make the images captures differently. Besides, as the spoofing displays can be produced using different kinds of materials, such as paper, electronic screen and forged mask, the distributions of spoofing samples are widely varied. Therefore, it is extremely hard that an approach trained on one dataset performs well on other datasets.

Among the numerous published literatures, most of them focused on hand-crafted features and tried to capture texture differences between live and spoofing face images from the perspective of surface reflection and material differences, such as LBP \cite{Chingovska2012On}, LBP-TOP \cite{Pereira2012LBP} and HOG \cite{komulainen2013context}. However, methods in this category may suffer from poor generalizability since the texture varies with the spoofing devices. In \cite{Yang2014Learn, li2016original}, the researchers used CNNs to automatically learn features for face anti-spoofing and have achieved promising performance. Nevertheless, the CNNs need a large number of various types of spoofing data to guarantee its generalization ability. Even the traditional approach for adapting deep models, fine-tuning, may require hundreds or thousands of labeled examples for each category that need to be adapted \cite{tzeng2015simultaneous}. Unfortunately, the current publicly available face spoofing datasets, such as Replay Attack \cite{Chingovska2012On}, CASIA-FASD \cite{Zhang2012A}, MSU-MFSD \cite{WenTIFS15} and NUAA \cite{tan2010face} are too limited to train a generalized network
compared with the datasets of image classification and face recognition \cite{tu2017illumination,luo2019learning,tu2019joint,tu2019joint,zhao2019multi}.

However, in the practical application of a face anti-spoofing product, it is reasonable to assume that the product's new owner will be able to label a handful of examples for a few types of training samples. Given this circumstance, we proposed a domain transfer network, aiming at learning domain-invariant features across two different datasets for robust face anti-spoofing. We propose to learn a shared feature subspace where the distributions of the real access samples (genuine) from different domains, and the distributions of different types of spoofing attacks (fake) from different domains
are drawn close, respectively. In the proposed framework, the sufficient labeled source data are used to learn discriminative representations that distinguish the genuine samples and the fake samples, meanwhile the sparsely labeled target samples are fed to the network to calculate the feature distribution distance between the genuine samples from the source and the target domain, and between the fake samples from the source and the target domains, corresponding to their materials. The kernel approach is adopted to map the features output from the CNN into a common kernel space, and the Maximum Mean Discrepancy (MMD) is adopted to measure the distribution distance between the samples from the source and target domains. This feature distribution distance is treated as a domain loss term added to the objective function and minimized along with training of the network. We provide a comprehensive evaluation on some popular datasets with the proposed method and show significant performance improvements.

\section{Related Work}

Existing face anti-spoofing approaches are mainly based on two cues for the purpose of liveness face detection, the texture differences between live and spoofing face images and the fine-grained motions such as eye blinking, mouth movement and head movement across video frames.

In \cite{li2004live}, the researchers utilized the difference of structural texture between the 2D images and 3D images to detect spoofing attacks based on the analysis of Fourier spectra, where the reflections of light on 2D and 3D surfaces result in different frequency distributions. Tan et al. \cite{tan2010face} used a variational retinex-based method and the Difference-of-Gaussian (DoG) filters to extract latent reflectance features on face images to distinguish fake images from real images. In \cite{maatta2011face}, Maatta et al. extracted the texture of 2D images using the multi-scale local
binary pattern (LBP) to generate a concatenated histogram which was fed into a SVM classifier for genuine/fake face classification. In the later work, Chingovska et al. \cite{Chingovska2012On} applied the LBP operator and its variation to capture the textural properties of the input image. However, these method analyze each frame in isolation, not considering the temporal motion cues across video frames. To make use of the temporal information, Pereira et al \cite{Pereira2012LBP}. proposed the LBP-TOP, considering three orthogonal planes intersecting the center of a pixel in the $XY$ direction,
$XT$ direction and $YT$ direction, where $T$ is the time axis. According to their experimental results, this multi-resolution LBP-TOP with SVM classifier achieved the best HTER of 7.6\% on the Replay-Attack dataset.
\begin{figure}[!t]
\centering{\includegraphics[width = 8.5cm, height=3.7cm]{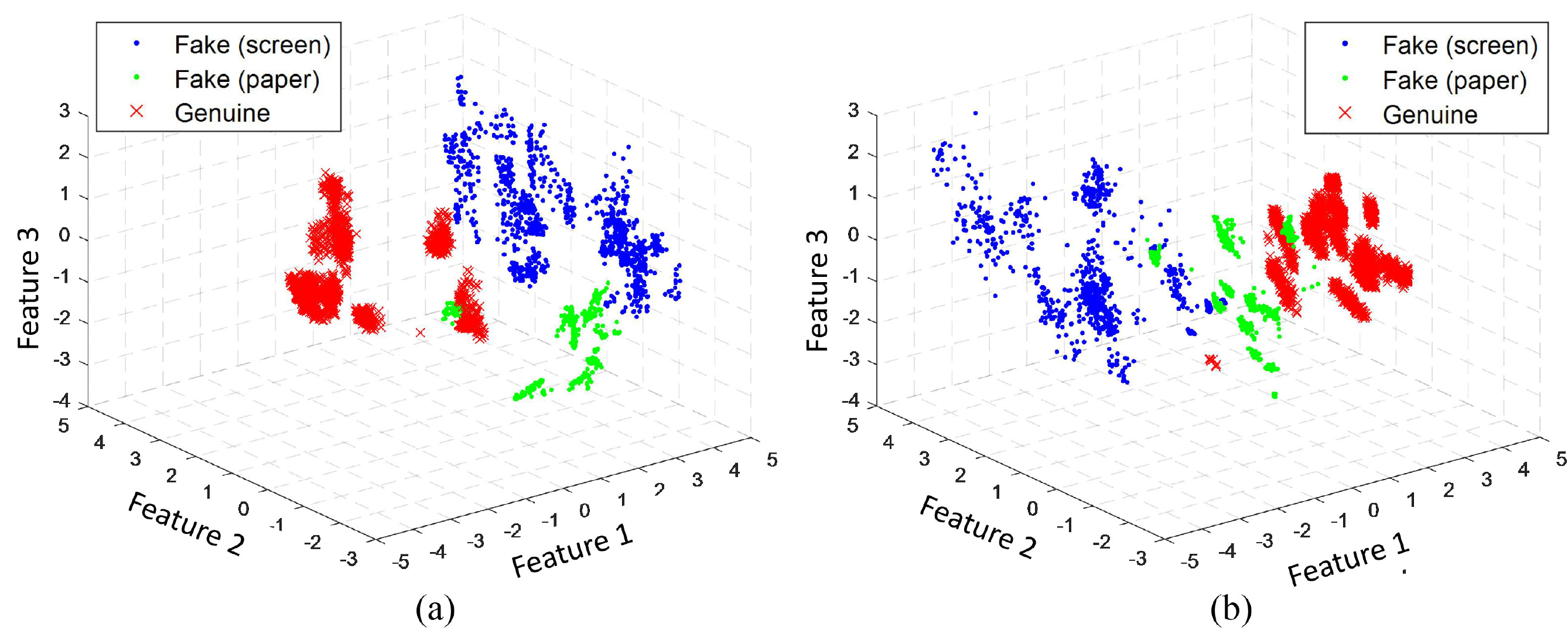}}
\caption{The distributions of genuine samples and two types of fake samples (print and video) from the database of MSU-MFSD (a) and Replay-Attack (b).
\label{fig1}}
\end{figure}

\begin{figure*}[!t]
\centering{\includegraphics[width = 16cm, height=5.0cm]{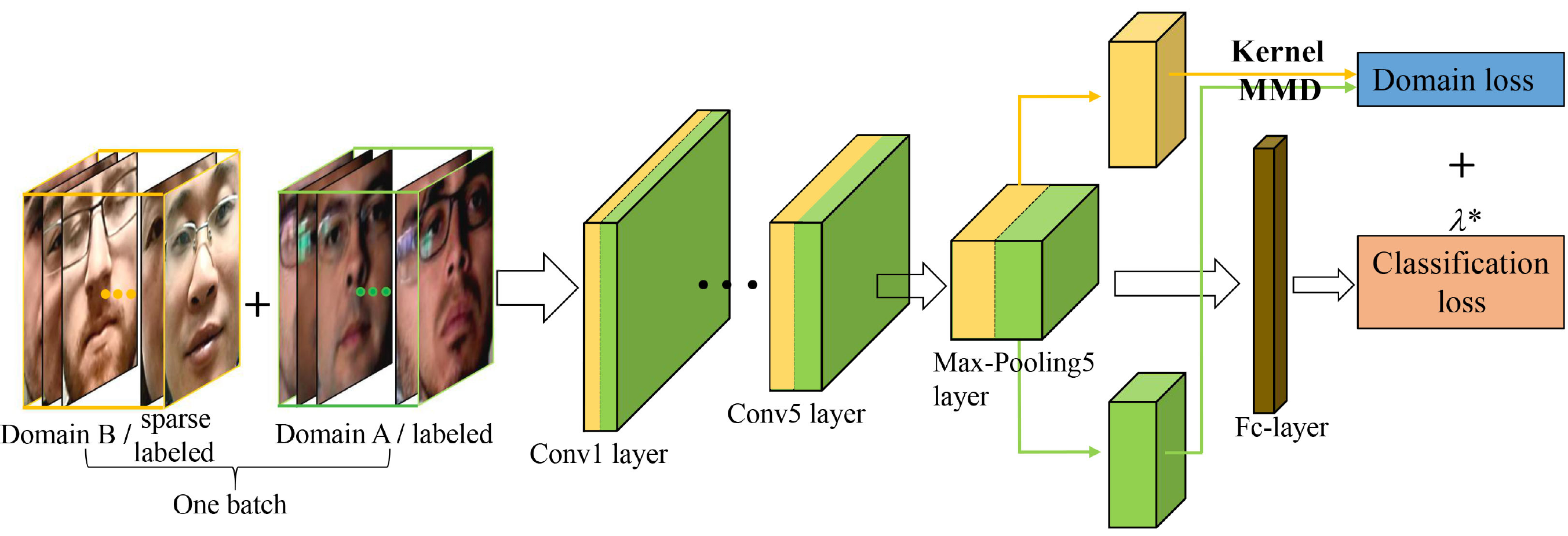}}
\caption{The flowchart of the proposed framework, where every input batch contains half the source images and half the target images. Features of the two domains output from the last pooling layer are used to calculate the distribution distance with kernel based MMD. The network is trained using the classification loss along with the distribution distance which is taken as domain loss.
\label{fig2}}
\end{figure*}

Though promising performance can be achieved with the above mentioned hand-crafted features in intra-dataset protocol, a new dataset of different domain may result in severe performance degradation, since many hand-crafted features are designed specifically and can not be easily transferred to new conditions \cite{bengio2013representation}. In order to obtain features with better generalization ability, quite a few researchers used the deep convolutional neural network (CNN) for face anti-spoofing \cite{Yang2014Learn,alotaibi2016deep,patel2016cross,valle2017transfer}. The CNNs can automatically learn features and obtain discriminative cues between genuine and fake faces. If sufficient data from various domains is available, the CNN-based methods can achieve performance with fairly good generalization ability. Unfortunately, the current publicly available face anti-spoofing datasets are too limited to train a generalized network due to the difficulty in obtaining labeled training samples, the variety of materials of spoofing devices also enhance the domain shift in the distribution of data representations.

To bridge the gap between two different distributions, many approaches have been proposed based on the ideas of domain
adaptation \cite{gopalan2011domain,fernando2013unsupervised,tzeng2014deep,ganin2016domain}. Multimodal deep learning architectures have been proposed to learn domain invariant representations in \cite{ngiam2011multimodal}. However, this method performed primarily in a generative context and did not leverage the full representation power of supervised CNN representations. In \cite{ghifary2014domain}, Ghifary et al. proposed pre-training with a denoising auto encoder, then train a two-layer network simultaneously with the MMD domain confusion loss. Nevertheless, the learned network is relatively shallow and therefore lacks the strong semantic representation which is learned by directly optimizing with a supervised deep CNN. In \cite{tzeng2014deep}, Tzeng et al. proposed a new CNN architecture which introduces an adaptation layer and an additional domain confusion loss for classification. They use MMD both to select the depth and width of the architecture while using it as a regularizer during fine-tuning, and achieved state-of-the-art performance on the standard visual domain adaptation benchmark.

\section{Deep Domain Transfer Network}
The idea of domain adaptation have been successfully applied in many other fields \cite{gopalan2011domain,chen2015deep}, similar ideas, however, have never been used in the study of face anti-spoofing, even current face anti-spoofing approaches severely suffered from the problem of dataset bias. In order to remove the dataset bias and bridge the gap between distributions from different domains with the limited training samples, we proposed a deep domain transfer framework for face anti-spoofing with kernel based metric. Kernel based metric has already been used in many works for the measurement
of distribution distance \cite{comaniciu2003kernel,xiong2014person,sutherland2016generative}. If a suitable kernel is found, the input data can be mapped into a convenient feature space, and the distance between different distributions will be better quantified.

We propose to use deep Convolutional Neural Network to extract the discriminative features of input images. We argue that the feature distributions of different datasets are different in feature subspace that learned by the shared feature extraction layer. Figure~\ref{fig1} illustrates the feature distributions of three principal components with respect to two different datasets, the features are learned by CNN hidden layer. It is shown that the distributions of the genuine samples and the two types of fake smaples in the learned feature subface are vary from dataset to dataset. The proposed framework is outlined in Figure~\ref{fig2}.

Let $x \in \mathbb{R}^d$ be a $d$-dimension column vector. $x^s \in \mathbb{R}^d$ and $x^t \in \mathbb{R}^d$ are the data points in the source and the target datasets, respectively. Suppose that the source-domain data $D^s = \left \{ (x_1^s, y_1^s), ..., (x_{n^s}^s, y_{n^s}^s) \right \}$ and the target-domain data $D^t =\\  \left \{ (x_1^t, y_1^t),..., (x_{n^t}^t, y_{n^t}^t) \right \}$, where $n^s$ and $n^t$ are the numbers of the data points. We denote the representation of the last pooling layer as $\phi(\cdot)$. Therefore, the representations of source-domain points, $x^s \in \mathbb{R}^d$, and target-domain points $x^t \in \mathbb{R}^d$, can be represented as $\phi(x^s)$ and $\phi(x^t)$, respectively. Suppose $K$ is the kernel of a reproduction Hilbert space $\mathcal{H}_k$ of functions. Then the MMD in $\mathcal{H}_k$ between the distributions of $P$ and $Q$ is \cite{gretton2012kernel}:
\begin{equation}
\begin{aligned}
\text{MMD}_K^2(P, Q) &:= \mathbb{E}_{x_p,x_p^{'}}[K(x_p, x_p^{'})] + \mathbb{E}_{x_q,x_q^{'}}[K(x_q, x_q^{'})] \\
&-2\mathbb{E}_{x_p,x_q}[K(x_p, x_q)]
\end{aligned}
\end{equation}
where $x_p, x_p'  \overset{iid}{\sim} P$ and $x_q, x_q'  \overset{iid}{\sim} Q$. Many kernels, such as the Gaussian RBF, are characteristic \cite{fukumizu2008kernel,sriperumbudur2010hilbert}, which indicates the MMD is a metric, and in particular that $\text{MMD}_K(P,Q) = 0$ if and only if $P = Q$. Given $X_p = \{X_p^1, ..., X_p^m\} \overset{iid}{\sim} P$ and
$X_q = \{X_q^1, ..., X_q^m\} \overset{iid}{\sim} Q$, one estimator of
$\text{MMD}_K(X_p, X_q)$ is:
\begin{equation}
\begin{aligned}
\hat{\text{MMD}_K}^2(X_p, X_q) &:= \frac{1}{m/2}\sum_{i \neq i^{'}}K(X_p^i, X_p^{i^{'}}) \\
&+ \frac{1}{m/2}\sum_{j \neq j^{'}}K(X_q^j, X_q^{j^{'}})\\
&- \frac{1}{m/2}\sum_{i \neq j^{'}}K(X_p^i, X_q^{j})
\end{aligned}
\end{equation}
This estimator is unbiased, and has nearly minimum variance among unbiased estimators \cite{gretton2012kernel}. Just take the feature representations $\phi(x^s)$ and $\phi(x^t)$ as $P$ and $Q$, respectively, the kernel based MMD between the features of source-domain and target-domain samples can be calculated accordingly. In our experiments, the mixture of RBF kernels were chosen for the computation of distribution distance based MMD.
\begin{table*}
\centering

{Table 1: The training, testing and development sets contained in databases of Replay-Attack and MSU-MFSD.}
\vspace{0.3cm}

\begin{tabular}{c|| c | c | c || c | c | c || c | c | c}
\Xhline{1.2pt}
     &     \multicolumn{6}{c||}{Replay-Attack-fixed}        &\multicolumn{3}{c}{{\multirow{2}*{MSU-MFSD}}} \\
\cline{2-7}
     &       \multicolumn{3}{c||}{Adverse}                  &\multicolumn{3}{c||}{Controlled} & \multicolumn{3}{c}{}            \\
\cline{2-10}
     &\multirow{2}*{Genuine} & \multicolumn{2}{c||}{Attack}&\multirow{2}*{Genuine}           & \multicolumn{2}{c||}{Attack} & \multirow{2}*{Genuine}&\multicolumn{2}{c}{Attack}  \\
\cline{3-4} \cline{6-7} \cline{9-10}
     &                        &    Video   &    Paper       &                                 &     Video      & Paper        &                       &   Video  & Paper \\
\Xhline{1.2pt}
Train&            30          &  30*4      &     30         &                 30              &      30*4      &      30      &        30             &   30*2   & 30  \\
\hline
Test &            40          &  40*4      &     40         &                 40              &      40*4      &      40      &        20             &   20*2   & 20  \\
\hline
Devel&            30          &  30*4      &     30         &                 30              &      30*4      &      30      &        20             &   20*2   & 20  \\
\Xhline{1.2pt}
\end{tabular}
\end{table*}

As Figure~\ref{fig2} shows, the proposed framework minimizes the distance between different domains, as well as the classification errors among genuine and fake samples. The features learned should be domain-invariant across domains to achieve good classification results on dataset of either the source domain or the target domain. If no labeled target domain data are available, the objection loss is defined as:
\begin{equation}
\begin{aligned}
\mathcal{L}_{uns} = \mathcal{L}_C(X_S, y) + \lambda \text{MMD}_K^2(X_S, X_T)
\end{aligned}
\end{equation}
where $X_S, X_T$ denotes the source-domain data and target-domain data, respectively, $y$ denotes the labels of source-domain data. $\mathcal{L}_C(X_S, y)$ is the classification loss on the labeled source-domain data, and $\text{MMD}_K^2(X_S, X_T)$ is the domain loss between the source data $X_S$ and target data $X_T$. The regulation parameter $\lambda$ determines how strongly we would like to confuse the domains.

In the situation of face anti-spoofing, spoofing images were usually obtained using different types of devices that have different materials of surfaces, i.e., print photo, electronic screen, forged mask, which we named spoofing modality in this paper. If sparsely labeled data from the target domain is available, we could define the objection function by a intra-modality way. Specifically, the source samples and target samples are divided into several subsets corresponding to the material of their spoofing surfaces (e.t, print paper, video screen, mask). Then, the MMD is calculated between different modalities. Therefore, the semi-supervised objection loss is:
\begin{equation}
\begin{aligned}
\mathcal{L}_{semis} &= \mathcal{L}_C(X_S, y) + \lambda \text{MMD}_K^2(X_S^{real}, X_T^{real}) \\
&+\lambda \sum_{i=1}^N \text{MMD}_K^2(X_S^{i}, X_T^{i})
\end{aligned}
\end{equation}
where $X_S^{real}$ and $X_T^{real}$ represent the real samples from the source and target domain, respectively, and $N$ is the number of spoofing modalities in the database. This objective function can be optimized by Adam algorithm \cite{kingma2014adam}. The Adam offers a computationally way for gradient-based optimization of stochastic objective functions, aiming at towards machine learning problems with large datasets and/or high-dimensional parameter spaces. It is robust and well-suited to a wide range of non-convex optimization problems \cite{kingma2014adam}.

We build the proposed Convnet based on AlexNet \cite{krizhevsky2012imagenet}, which contains five convolutional and max pooling layers, and three fully connected layers. A "Two-Half" strategy is proposed for the training of the proposed framework. Specifically, in every training batch, one half is the source data and the other half is the target data, as illustrated in Figure~\ref{fig2}. Since only a few target samples are available, we randomly picked and copied these target samples to ensure the two domains have the same number of samples. During the training process, the labeled
samples from the source domain are used to learn discriminative features for the distinguish of genuine and fake images, while the target samples are used to shrink the domain variance against the source domain, corresponding to the modalities, respectively. Owning to the two terms of joint loss, the proposed deep domain adaptation network could learn representations that are effectively discriminative between genuine and fake samples due to the classification loss, while still remaining invariant to domain shift due to the domain loss.
\begin{table*}
\centering

{Table 2: Comparison results of inter-test between different pairs of datasets. The training set of the first column are used to train the model,
    whereas the testing set of the second column are used for model testing.}
\vspace{0.3cm}

\begin{tabular}{c| c || c | c | c | c | c }
\Xhline{1.2pt}
\multicolumn{2}{c||}{\multirow{2}*{\large{Datasets}}}   & \multirow{2}*{\large{Metrics}} & \multicolumn{4}{c}{\multirow{2}*{\large{Methods}}} \\
\multicolumn{2}{c||}{}                         &                 & \multicolumn{2}{c}{}                       \\
\Xhline{1.2pt}
\multirow{2}*{Train}     & \multirow{2}*{Test} & \multirow{2}*{} & \multirow{2}*{$LBP_{8,1}^{u2}$} & \multirow{2}*{$LBP$-$TOP_{8,8,8,1,1,1}^{u2}$} & \multirow{2}*{$DoG$} & \multirow{2}*{$DTCNN(proposed)$} \\ [-0.0ex]
               &               &              &  &  &  & \\
\hline
Replay-attack  & Replay-attack & HTER$_e$(\%) & 22.50 & 24.00 & 24.42 &\textbf{20.00} \\
(adverse)      &(controlled)   & AUC(\%)      & 77.50 & 76.00 & 75.58 &\textbf{80.00} \\
\hline
Replay-attack  &   MSU-        & HTER$_e$(\%) & 45.83 & 47.50 & 36.36 &\textbf{25.83}     \\ [0.3ex]
(adverse)      &     MFSD      & AUC(\%)      & 54.17 & 52.50 & 63.64 &\textbf{74.17}     \\ [0.3ex]
\hline
MSU-           & Replay-attack & HTER$_e$(\%) & 45.50 & 46.50 & 48.58 &\textbf{27.50} \\ [0.3ex]
MFSD           &(controlled)   & AUC(\%)      & 54.50 & 53.50 & 51.42 &\textbf{72.50} \\ [0.3ex]
\Xhline{1.2pt}
\end{tabular}
\end{table*}

\section{Experiments}
In this section, we first give a brief description of two benchmark datasets, Replay-Attack \cite{Chingovska2012On} and MSU-MFSD \cite{WenTIFS15}. After that, we report the performance evaluation of the proposed method on the two datasets. We follow the protocols provided by these two datasets.

\subsection{Experimental Settings and Datasets}
To avoid any influences from the background, only the face region of each image was cropped based on eye coordinates and used as input. The eye coordinates were obtained by a facial landmarks detection algorithm available on the internet. All the input face images were resized to $224 \times 224$, and the regulation parameter $\lambda$ was set as 0.5. Batch normalization layer was used to overcome internal covariate shift. The mixture of Gaussian RBF kernels were chosen for the calculation of the domain loss. The mixed kernel function is a sum of Gaussian RBF kernels with fixed bandwidths 2, 5, 10, 20, 40 and 80.

MSU-MFSD: This dataset consists of 280 video recordings of genuine and attack faces. The recordings were taken from 35 individuals using two types of cameras, with different resolutions ($640 \times 480$ and $720 \times 480$). For the real accesses, each subject has two video recordings captured with the Android and Laptop cameras. For the video attacks, a high definition video was taken for each subject using a Canon camera and a iPhone camera. The videos taken with the Canon camera were then replayed on iPad Air screen to generate the HD replay attacks while the videos recorded by the iPhone mobile were replayed itself to generate the mobile replay attacks. For the printed attacks, the pictures were printed on A3 Paper using an HP colour printer. The 35 subjects (280 videos) of the MSU-MFSD dataset were divided into training (120 videos) and testing (160 videos) subdatasets, respectively. The training dataset contains 30 real access videos and 90 attack videos while the testing dataset contains 40 real accesses and 120 attacks.

Replay-Attack:  This dataset consists of 1200 videos that include 200 real access videos and 1000 attack videos. The attacks were taken under 2 different illumination conditions (controlled and adverse) and 2 support conditions (hand and fixed). Under the same condition, a high resolution picture and video were taken for each person. Three types of attacks were designed: (1) \textsl{print attack}, (2) \textsl{mobile attack}, and (3) \textsl{highdef (high definition) attack}. The evaluation protocol splits the dataset into training (360 videos), testing (480 videos) and development
(360 videos) subdatasets. The training and development datasets contain 60 real access videos and 300 attack videos each, whereas the testing subdataset contains 80 real accesses and 400 attacks. In our experimetns, only the attacks that using fixed-support were used.

\subsection{Performance measure}
We evaluate the adaptation ability of the proposed transfer network in this section. Half Total Error Rate (HTER) was used as the metric in our experiments to keep consistent to previous work. The HTER is half of the sum of the False Rejection Rate (FRR) and the False Acceptance Rate (FAR). Since both False Acceptance Rate (FAR) and False Rejection Rate (FRR) depend on a threshold $\tau$, increasing the FAR will usually reduce the FRR and vice-versa, we followed previous works \cite{Yang2014Learn,xu2015learning}, and used the development set to determine the threshold $\tau$ corresponding to Equal Error Rate (ERR) for the computing of HTER. Unfortunately, there is no development set in MSU-MFSD dataset. In this case, we equally split the testing set into a couple, namely the development set and the new testing set. The split datasets are shown in table 1. Observing from table 1, there are 30, 40 and 30 subjects used for training, testing an development in Replay-Attack database, respectively, while the training, testing and development datasets in MSU-MFSD each contains 30, 20 and 20 subjects, respectively. The testing accuracy (ACC) achieved by each methods is also calculated for comparison.

\subsection{Cross-Databases Spoofing Detection}
For face anti-spoofing, the adaptation ability from one dataset to another is crucial for practical application. In this part, we evaluate this ability by cross-dataset testing (inter-test), namely the model is firstly trained using the samples from dataset A (source), and then tested on dataset B (target). The Replay-Attack and the MSU-MFSD databases were used to evaluate the proposed method. Images in Replay-Attack were divided into two subdatasets according to their illumination conditions (adverse or controlled), and then, we could obtain three datasets: MSU-MFSD, Replay-Attack-controlled and Replay-Attack-adverse. Replay-Attack-controlled is the subdataset of Replay-Attack where images are taken with the light controlled, whereas Replay-Attack-adverse is the other subdataset of Replay-Attack with the light uncontrolled. For each of the three datasets, the images were split into three subsets, the training, testing and development set. To execute the experiments in inter-test fashion, the training set of dataset A was used to train the CNN models or train the SVM classifiers, whereas the testing set from dataset B was used for testing. Three groups of inter-test performance were evaluated: Repaly-Attack-adverse vs Replay-Attack-controlled, Repaly-Attack adverse vs MSU-MFSD, and MSU-MFSD vs Replay-Attack-controlled.

We consider the situation that the labeled data from target domain is limited for training. In our experiments, we only randomly selected one labeled subject from the target dataset for training. The video frames, including genuine and fake, were stacked and copied to the same number with the training samples from source domain. Because the input data were video clips, each contained a number of consecutive frames, we took all the consecutive frames for the training of the network for the purpose of data argumentation. However, in the testing phase, the output probabilities of the consecutive frames with respect to the same subject were averaged to determine the categorization of each video clip. Three popular methods, $LBP_{8, 1}^{u2}$ \cite{Chingovska2012On}, $LBP$-$TOP_{8,8,8,1,1,1,}^{u2}$ \cite{Pereira2012LBP} and $DoG$ \cite{Zhang2012A}, were implemented to compare with the proposed method. The features captured by the three methods are all fed into SVM to obtain the final classification results. For the LBP method, all the consecutive frames of each video clip of the training set were used to extract the 59-dimensional holistic features, and  to train the SVM classifier. The final results were achieved on the test set by averaging the probabilities of all the consecutive frames per video clip. The features of LBP-TOP were extracted per video clip, as this method involved the spatio-temporal information from video sequences. For the Dog operator, 30 frames of each video clip were randomly selected to train the SVM classifier to avoid high dimensionality of the DoG features, followed by the procedures of the original literature \cite{Zhang2012A}. The standard CNN (stdCNN) was also implemented to compared with the three methods, the stdCNN has the same architecture with the proposed framework only without the domain loss layer.

The inter-test results regarding these methods are first summarized in table 2 to evaluate their generalization ability on the three datasets. For Replay-Attack-adverse and Replay-Attack-controlled datasets, only the illumination condition is different, other factors like individuals and spoofing materials are keep unchanged, therefore the domain shift is not serious between these two datasets. This is the reason why all the four approaches achieved the best performance by testing across these two datasets. However, the MSU-MSFD database is totally different from Replay-Attack database, where the individuals, spoofing materials and illumination conditions are all different. As a result, the domain shift between these two databases are quite serious. The performances achieved by the four approaches are relatively lower across there two databases.

\begin{figure*}[!t]
\centering{\includegraphics[width = 17.5cm, height=15cm]{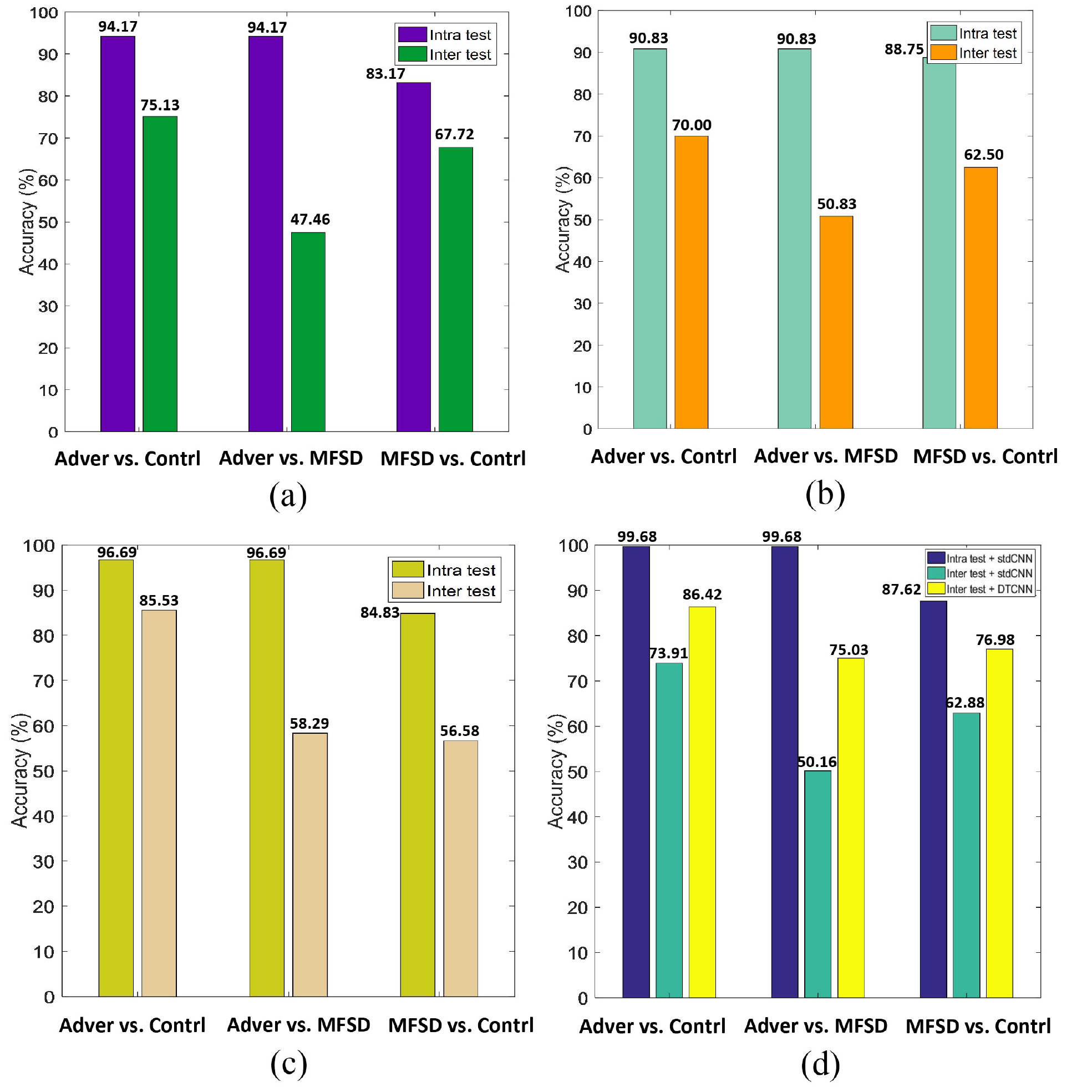}}
\caption{The accuracy of intra- and inter test achieved by LBP (a), LBP-TOP (b), DoG (c) and stdCNN vs. DTCNN (d), where stdCNN stands for the standard CNN, and DTCNN represents for the domain transferred CNN.
\label{fig3}}
\end{figure*}

\subsection{Intra-test vs. Inter test}
To illustrate the effectiveness of the proposed method, we compared the intra-test performance with inter-test performance regarding the same method, LBP, LBP-TOP, DoG and stdCNN. Suppose we evaluate the performance across dataset A and dataset B, then the intra-test was executed by using the training and testing set from A for the training and testing of the model. However, for the inter-test procedure, we used the training set from A and the testing set from B to train and test the model, respectively.

Figure~\ref{fig3} reports the accuracy of intra-test and inter-test by LBP, LBP-TOP, DoG, stdCNN and the proposed method (DTCNN), respectively. As can be seen, all of the methods obtained satisfying performance in the fashion of intra-test, the CNN even achieved nearly perfect performance ($99.68\%$) on Replay-Attack database. However, when intra-test on MSU-MFSD, we observed the training accuracy is nearly $100\%$, while the testing accuracy degraded significantly to $87.62\%$. The main reason for such a performance decline is over-fitting due to short of training samples.

The best inter-test performances were still achieved when testing across Replay-Attack-adverse and Replay-Attack-controlled (Adver vs. Contrl). Compared with the results of intra-test, there was a significant degeneration of the performance of inter-test by the standard CNN (blue vs. green bar graph in Figure~\ref{fig3}(d)). The declined accuracy even reached up to $49.52\%$ when testing across Replay-Attack-adverse and MSU-MFSD (Adver vs. MFSD). However, when using the proposed method (DTCNN) for inter-test (blue vs. yellow bar graph in Figure~\ref{fig3}(d)), there was a considerably boost of the performance compared with the standard CNN. The improved accuracy are $12.51\%, 24.87\%$ and $14.1\%$ for Adver vs. Contrl, Adver vs. MFSD and MFSD vs. Contrl, respectively.

Overall, the results of LBP, LBP-TOP, DoG and the standard CNN show that they are not able to well handle the cross-database challenge. This is because LBP, LBP-TOP and DoG are hand-crafted features which are designed specifically and can not be easily transferred to new condition. For the standard CNNs, the features learned are specific to the training dataset provided, it may achieve superb performance on the data from the training domain, however, the performance may degrade considerably if testing on a total different dataset. Different from the standard CNNs, the proposed method can make use of the domain information from the target dataset, bridging the gap between the feature distributions and hence obtaining satisfying performance on the target set, only with a very few labeled samples provided from the target domain.

\section{Summary and Conclusion}
Cross-database face anti-spoofing replicates real application scenarios, and is a challenge for biometrics anti-spoofing. Although many of the existing methods proposed can achieve excellent performance in the way of non-realistic intra-database testing, few of them can achieve comparable performance on the dataset of other domain. To bridge the gap between the datasets from different domains, we proposed a CNN framework that effectively adapts to a new domain with sparsely labeled target domain data for face anti-spoofing. The proposed network can learn a invariant feature space for the source and target samples by optimizing an objective that simultaneously minimizes classification loss and the domain loss. As a result, the model trained with the labeled source data can also achieve satisfactory performance on the target dataset. Experiments on the datasets of Replay-Attack and MSU-MFSD showed the proposed framework greatly enhance the performance by cross-dataset testing, with only a few labeled samples from the target domain. The proposed method could open new perspectives for the future research of face anti-spoofing.

{\small
\bibliographystyle{ieee}
\bibliography{mybibfile}
}

\end{document}